\documentclass{article}
\usepackage[utf8]{inputenc}

\usepackage{fullpage}
\usepackage{array}
\usepackage{caption}
\usepackage{subcaption}
\usepackage{graphicx}
\usepackage{hyperref}
\usepackage{multirow}
\usepackage{amsfonts}
\usepackage{amsmath}
\usepackage{geometry}
\usepackage{amssymb}
\usepackage{float}
\usepackage{tikz}
\usetikzlibrary{mindmap}
\usetikzlibrary{shapes, snakes, arrows}
\usetikzlibrary{positioning,fit,calc}
\usepackage{amsthm}
\usepackage{comment}
\usepackage{authblk}
\usepackage{booktabs}

\theoremstyle{plain}
\newtheorem{theorem}{Theorem}[section]
\newtheorem{proposition}[theorem]{Proposition}
\newtheorem{lemma}[theorem]{Lemma}

\theoremstyle{definition}
\newtheorem{definition}[theorem]{Definition}

\theoremstyle{remark}
\newtheorem{remark}[theorem]{Remark}

\tikzset{block/.style={draw, thick, text width=3cm, minimum height=1cm, align=center},   
line/.style={-latex}   
}  

\tikzset{elips/.style={ellipse,draw,minimum width=4em,minimum height=1cm, align=center, inner ysep=0pt}}

\usepackage[
    backend=biber,
    style=numeric,
    hyperref = true,
    natbib = true, 
    maxbibnames=9,
    maxcitenames=1
  ]{biblatex}

\title{Uncertainty in Fairness Assessment: Maintaining Stable Conclusions Despite Fluctuations}

\author[1,2]{Ainhize Barrainkua}
\author[1,3]{Paula Gordaliza}
\author[1,4]{Jose A. Lozano}
\author[1,2,5]{Novi Quadrianto}

\affil[1]{Basque Center for Applied Mathematics (BCAM), Spain}
\affil[2]{Predictive Analytics Lab (PAL), University of Sussex, UK}
\affil[3]{Universidad Pública de Navarra (UPNA), Spain}
\affil[4]{University of the Basque Country (UPV/EHU), Spain}
\affil[5]{Monash University, Indonesia}

\begin{document}

\maketitle

\begin{abstract}
   Several recent works encourage the use of a Bayesian framework when assessing performance and fairness metrics of a classification algorithm in a supervised setting. We propose the \textit{Uncertainty Matters} (UM) framework that generalizes a Beta-Binomial approach to derive the posterior distribution of any criteria combination, allowing stable performance assessment in a bias-aware setting.
   We suggest modeling the confusion matrix of each demographic group using a Multinomial distribution updated through a Bayesian procedure. We extend UM to be applicable under the popular $K$-fold cross-validation procedure.
   Experiments highlight the benefits of UM over classical evaluation frameworks regarding informativeness and stability.   
\end{abstract}

\section{Introduction}
\label{sec:intro}

With the current adoption of machine learning (ML) systems in social, economic, and industrial domains, concerns about the fairness of automated decisions have been added to the problem of ensuring the efficiency of algorithms in a stable and interpretative manner. Although both aspects are measured in terms of performance metrics, fairness entails the additional challenge of incorporating sensitive information in the data and new procedures need to be considered to control the stability of such outcomes.

\begin{figure}
    \centering
    \includegraphics[width=0.6\textwidth]{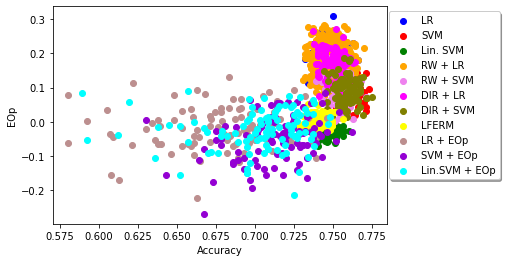}
    \caption{10-fold CV results regarding accuracy and fairness, measured by means of Equality of opportunity (EOp), for 10,000 different 10-fold CV configurations. Each point refers to one 10-fold CV result, and the colors refer to different benchmark fairness-enhancing methods. These results were obtained using the German dataset with age as the sensitive attribute.}
    \label{fig:german_age_av}
\end{figure}

Recent ML trends are increasingly encouraging researchers to incorporate uncertainty into the evaluation of algorithm-based systems.
 In order to increase the transparency of algorithmic performance measures, typically for comparison purposes, some authors \cite{benavoli2017time, kruschke2018bayesian} propose to treat these metrics as random variables whose posterior distributions are updated through Bayesian inference. In the fair learning setting, these kinds of considerations are also necessary, especially since fairness metrics have been proved unstable with respect to dataset composition. In particular, \citet{ji2020can} or \citet{friedler2019comparative} showed how certain fairness metrics strongly vary, respectively, in hold-out evaluations or under different training-test splits in the datasets typically used in the algorithmic fairness literature. This is also observable through other popular evaluation methods such as 10-fold cross validation (CV) in Figure \ref{fig:german_age_av}, and more so for fairness metrics than for performance metrics.

Several works have studied the problem of accurately assessing the uncertainty of fairness metrics in a supervised scenario. Early proposals providing a probabilistic approach focused on a sole fairness metric, usually Demographic Parity (DP). In \citet{besse2018confidence, besse2021survey} the asymptotic distribution of Disparate Impact, one of the most
classic indexes for quantifying DP, was obtained to build confidence intervals using the traditional Delta method. Later, \citet{ji2020can} considered a Bayesian framework and a calibration procedure to reduce such uncertainty by employing unlabeled data. However, existing works on this topic have generally in common two main limitations. First, the simultaneous comparison of several performance and fairness metrics is not possible. Second, metric uncertainty can only be addressed for already trained predictors or learning algorithms under the hold-out evaluation, but not for more complex and popular evaluation frameworks, such as $K$-fold CV, where the $K$ results obtained are not independent 
due to the overlap of training instances.

In order to address these shortcomings, we propose \textit{Uncertainty Matters (UM)}, a new framework based on Bayesian inference to account for the uncertainty inherent in the reported metrics by modeling the confusion matrix (CM) itself. Particularly, performance and fairness metrics are assumed to be random variables whose distribution is updated by Bayesian inference with the results obtained after performing a classification task. The main advantage of our formulation is that it allows the straightforward computation of any $N$-dimensional joint posterior distribution. 
Importantly, we extend the statistical framework to the well known $K$-fold CV evaluation scenario, allowing to derive the uncertainty of performance and fairness guarantees of learning algorithms based on a so-called \textit{effective CM} \cite{wang2019bayes}. In addition, we design a scheme for algorithmic comparison based on the UM framework that aims at analyzing the gap existing between two algorithms with respect to an extended notion of the Region of Practical Equivalence \cite{benavoli2017time} to general dimension.


We have carried out extensive experiments over several benchmark datasets and fairness-enhancing interventions which have highlighted the benefits of our UM framework over classic evaluation settings. 
It broadens the perspective regarding the classifiers behavior, providing more insights; and, importantly, the results are stable w.r.t. dataset composition, enabling us to derive more trustworthy conclusions.





The rest of the paper is organized as follows. 
We describe our statistical framework UM in Section \ref{sec:uncertainty} and propose a strategy based on it for algorithmic comparison in Section \ref{sec:comparison}.
Section \ref{sec:related} provides an overview about existing works linked to our approach. 
Finally, Section \ref{sec:experiments} describes the experiments for the empirical evaluation of our proposal and the main results derived from it. The main conclusions are summarized in Section \ref{sec:conclusion}, where we also identify future directions.

\section{The UM framework for the Uncertainty-Aware Assessment of Fairness and Performance Guarantees}
\label{sec:uncertainty}

The Beta-Binomial approach has been widely used to describe the marginal distributions of performance and fairness metrics \cite{ji2020can, goutte2005probabilistic}.
However, the validation of fair interventions requires 
simultaneous comparison of multiple targets expressing performance and fairness guarantees. 
It is therefore essential to obtain the joint distributions of performance (e.g., accuracy) and fairness 
(e.g., parity in true positive rate) metrics, which is not possible using the Beta-Binomial approach since the dependence needs to be characterized. In this section, we propose the UM framework that models the uncertainty related to the CM, and allows to hierarchically conclude the multi-dimensional posterior distributions.

\subsection{Notation and Problem Statement}
\label{sec:notation}


Our approach is built upon a bias-aware supervised learning scenario: let $\mathcal{M}$ be the predictive algorithm trained using a finite set of examples i.i.d. from $P(\mathcal{X} \times \mathcal{S} \times \mathcal{Y})$, where every instance is represented by a set of $d$ (non-sensitive) attributes $\textbf{x} \in \mathcal{X} \subset \mathbb{R}^d$, a sensitive attribute $s \in \mathcal{S}$ (e.g., gender, age, marital status) and a class label $y \in \mathcal{Y}$. For explanatory purposes, we will refer to the binary classification problem where $\mathcal{Y}= \{ -1, +1 \}$, but UM works for multiclass $\mathcal{Y}$. We denote by $CM_s$ the confusion matrix of each sensitive group $s$, consisting of the number of true positive $TP_s$, true negative $TN_s$, false positive $FP_s$ and false negative $FN_s$. 

If $P(\mathcal{X} \times \mathcal{S} \times \mathcal{Y})$ were explicitly known, the performance and fairness metrics of any algorithm $\mathcal{M}$ could be evaluated precisely. However, generally, only an approximation of the theoretical values can be obtained from a set of finite samples i.i.d. from such distribution.
Therefore, we propose to view fairness and performance metrics as random variables and model the uncertainty inherent in the evaluation process to, accordingly, report trustworthy information regarding the behavior of the classifier on the whole population.


Let $\mathcal{T}$ be the set of performance metrics. We realize that, in a classification problem, any $\Theta \in \mathcal{T}$ could be obtained from the CM, $\Theta=f(CM)$. Moreover,
it is important to note that fairness is generally defined in terms of the values of certain performance metrics restricted to different sensitive groups. More precisely, a classifier is said to be (almost) fair w.r.t. $\Theta \in \mathcal{T}$ if $\Theta$ only changes slightly when restricted to samples from different subgroups $s\in \mathcal{S}$. Therefore, to avoid potential ambiguity, fairness metrics will be denoted by $\eta$ and we will consider them as functions of the CM in the sensitive groups $\eta=g(CM_s)$. For instance, in the particular case of $\mathcal{S}=\{0,1\}$, 
a one-dimensional metric such as DP would be measured by $\eta=AR_1-AR_0$ (difference in acceptance rate across groups); while for a multidimensional one, such as Equalized Odds (EO), would be quantified as $\boldsymbol\eta=(TPR_1-TPR_0,FPR_1-FPR_0)$, (difference in true positive rate and in false positive rate across groups, respectively).
See \citet{verma2018fairness} for an extensive review on fairness metrics.

\subsection{Probabilistic Model}

In order to account for multiple behavioral guarantees of an algorithm, we propose a procedure, called \textit{Uncertainty Matters} (UM), to derive the joint posterior distribution $P(\boldsymbol\Theta, \boldsymbol\eta)$ of a sequence of user-defined metrics of interest $ (\boldsymbol\Theta, \boldsymbol\eta)= (\Theta_1,\ldots, \Theta_N ,\eta_1,\ldots, \eta_M ), $ where $ \boldsymbol\Theta \subset \mathcal{T}$ and, for every $i\in\{1,\ldots,N\}$ and $j\in\{1,\ldots,M\}$, there exist functions $f_i$ such that $\Theta_i=f_i(CM), \ i=1,\ldots, N,$ and functions $g_j$ such that $\eta_j=g_j(CM_s), \ j=1,\ldots,M$

 
Yet, in general, a closed form for $P(\boldsymbol\Theta,\boldsymbol\eta)$ is very difficult to obtain, and instead we propose to compute its empirical counterpart on a set 
of $T$ samples 
$P_n(\{ (\Theta_1^t,\ldots, \Theta_N^t, \eta_1^t,\ldots, \eta_M^t) \}_{t=1}^T)$, obtained through a posterior hierarchical sampling procedure outlined in \eqref{eq:sampling}.



\begin{remark} While modeling the $N$-dimensional distributions in close form is generally a complex task, there are certain situations where it could be simplified. In particular, if the definitions of the considered metrics are given in terms of different counts of the CM, the joint distribution can be decomposed into the product of the marginal distributions of such metrics.
For example, the posterior distribution of the 2-dimensional notion EO 
can be computed as the product of the marginal distributions of $\Delta TPR$ and $\Delta FPR$.   
\end{remark}
  
The UM approach is grounded in the probabilistic modeling of the CM, from which the estimation of performance metrics $\boldsymbol\Theta \subset \mathcal{T}$ could be derived. Following \citet{caelen2017bayesian}, we propose to adopt a Bayesian framework on the CM, assuming that its values are drawn from a multinomial distribution. More precisely, for a bias-aware analysis, if we denote by $\mathcal{D}_s=(TP_s, TN_s, FP_s, FN_s)$ the results on independent samples of each sensitive group $s$, we write
\begin{equation*}
     \textrm{CM}_s \equiv \mathcal{D}_s \sim \textrm{MLT}(N_s;\pi_s),
\end{equation*}
where $N_s = TP_s + TN_s + FP_s + FN_s$, and the multinomial parameter $\pi_s = (\pi_{TP_s}, \pi_{TN_s}, \pi_{FP_s}, \pi_{FN_s}) \in \left[0,1\right]^4$ is assumed to follow a Dirichlet distribution, as it is the conjugate of the multinomial distribution. Hence, starting from the \textit{ prior} distribution $\pi_s \sim \textrm{Dir} (\alpha_s) = \textrm{Dir} ((\alpha_1^s, \alpha_2^s, \alpha_3^s, \alpha_4^s))$, by Bayesian inference, the \textit{posterior} distribution of the multinomial parameter will be:
\begin{equation}
 \pi_s | \mathcal{D}_s \sim \textrm{Dir} ((\alpha_1^s + TP_s, \alpha_2^s + TN_s, \alpha_3^s+FP_s, \alpha_4^s+FN_s))
    \label{eq:multinomialparameter}
\end{equation}

That is, for each sensitive group $s$, the CM$_s$ follows a multinomial distribution whose parameter $\pi_s$ is updated by $\mathcal{D}_s$. This would be the basic tool from which we would be able to derive joint posterior distributions of performance and fairness metrics, and eventually design uncertainty-aware procedures to evaluate algorithms. We will consider two distinct evaluation frameworks: Hold-out in Section \ref{sec:hold_out} and $K$-fold CV in Section \ref{sec:k_fold}.



\subsection{Hold-out Evaluation}
\label{sec:hold_out}

One of the most intuitive ways of evaluating a learning algorithm is the hold-out procedure: (randomly) splitting the data into two independent sets, using the former to train the algorithm, and the latter, for evaluation purposes. 
In this section, we design a hierarchical procedure to assess the uncertainty-aware fairness and performance guarantees of a learning algorithm on an independent sample of instances, assuming the above probabilistic model for the CM.



Based on the predictions of the trained model on the test instances for each group $s$, the confusion matrix $\hat{CM}_s$ is obtained. Then, from \eqref{eq:multinomialparameter}, we simulate for every sample $t=1,\ldots,T$ the value (i) $\pi_s^t$ of the parameter of the Dirichlet distribution, 
  from which (ii) the posterior multinomial distribution of $CM_s^t$ is derived.  As a result, (iii) the values of any performance metric $\Theta_{i}^t$ and any fairness metric $\eta_j^t$ could be computed as a function of $CM_s^t$. Consequently, we achieve the approximation
$P(\boldsymbol\Theta,\boldsymbol\eta) \approx P_n(\{ (\Theta_1^t,\ldots, \Theta_N^t, \eta_1^t,\ldots, \eta_M^t) \}_{t=1}^T)$.
\vspace{-0.15cm}
\begin{equation}
    \begin{cases}
      \text{(i)} \; \; \; \; \; \pi_s^t \sim P(\pi_s | \hat{CM}_s)\\
     \text{(ii)} \; \; \; \; CM_s^t \sim P(CM^s |\pi_s^t) \\
    \text{(iii)} \; \; \; \Theta_{i}^t = f_i(CM_{s}^t), \; i = 1,\ldots, N \\
    \quad \quad \; \; \eta_j^t = g_j(CM_{s}^t),  \;  j = 1,\ldots, M \\
    \end{cases}
     \label{eq:sampling}
\end{equation}

Note that \eqref{eq:sampling} is also applicable whenever the learning algorithm is trained on a dataset with an specific train/test partition (e.g. Adult Income), or when we have an already trained (possibly black-box) model. 

In the hold-out evaluation scenario, only a single train/test split of a dataset is considered. Yet, in practice, most works in the literature adopt alternative evaluation scenarios with multiple train/test splits of the same dataset, making it more difficult to derive trustworthy uncertainty estimates. This is particularly the case of $K$-fold CV, which has been widely used for comparative purposes in the fair learning setting. The next section describes the proposed approach to derive uncertainty-aware results of a $K$-fold CV.

\subsection{K-fold Cross-Validation}
\label{sec:k_fold}

The $K$-fold CV is a popular evaluation framework for limited data scenarios based on a resampling procedure. For our particular purpose, $K$ different posterior distributions of fairness and performance metrics could be concluded from the CM obtained in each fold following procedure \eqref{eq:sampling}. Nonetheless, the final goal is to obtain a single joint posterior distribution of such metrics which describes the guarantees of the learning algorithm. In this sense, it is important to note that the preliminary idea of considering a linear combination of the $K$ posterior distributions is not straightforward since they are correlated. Consequently, accounting for such a correlation is essential to accurately estimating the posterior distribution of the performance of the learning algorithm.
In the next section, we describe how to account for this with a so-called \emph{effective} confusion matrix $CM_s^e$.
The elements of this matrix represent the number of independent observations that are equivalent to the correlated solutions obtained from the $K$-fold CV. 



\subsubsection{Effective confusion matrix}

Let us denote by $CM_s^{(k)}$ the confusion matrix obtained for each sensitive group $s$ in the fold $k=1,...,K$.
We propose to approximate the information provided by the sequence of $K$ confusion matrices $\{CM_s^{(k)}\}_{k=1,\ldots,K}$ through a single matrix $CM_s^{e}$, namely the \textit{effective} confusion matrix \cite{wang2019bayes}. 
The latter is described by the number of independent observations that are equivalent to the correlated solutions obtained from, in our case, the $K$-fold CV. 
The following proposition describes the form of an \textit{effective} CM for $K$-fold CV (we omit $s$ for clarity but it also holds for $CM_s$).


\begin{proposition}
The effective confusion matrix for a $K$-fold CV is: 
\begin{equation}
    CM^e =  \frac{1}{1+ (K-1)\rho} \sum_{k=1}^K CM^{(k)},
    \label{eq:effective_k}
\end{equation}
where $ CM^{(k)}$ is the CM obtained in the $k$-th train/test split configuration and $\rho$ denotes the correlation between the results obtained for every train/test configuration in one $K$-fold cross-validation. 

\label{prop:effectivecm}
\end{proposition}
  The proof of Proposition \ref{prop:effectivecm} can be found in Appendix \ref{sec:proofs}.
  Note that for $\rho=0$ and $K=1$ we obtain the results for the hold-out evaluation. Otherwise, the true value of the correlation $\rho$ is unknown and varies for different problems. Therefore, estimating such quantity $\rho$ is one of the main challenges of this approach. 
If we manage to do this, the variance of performance metrics that are computed based on effective confusion matrix will not be underestimated and the joint posterior distributions could be computed for the K-fold CV results following a hierarchical procedure analogous to that described in Section \ref{sec:hold_out} for the hold-out evaluation. The only difference is that the distribution of the multinomial parameter for each sensitive group $s$ would be updated by the effective confusion matrix $CM_s^e$.

Let us now return to the estimation of the correlation $\rho$. 
Several approaches have already been proposed in the literature. 
For instance, \citet{nadeau2003inference} propose to approximate it as $\rho_0 = 1/K$, which is accurate when the Vapnik–Chervonenkis (VC) dimension of the algorithms is not too large compared to the size of the training set (e.g., a parametric model that is not too complex), or for algorithms that are robust to perturbations in the training set (e.g. SVM). Other works, such as the one by \citet{wang2019bayes} assume that $\rho \in [0, 1/K]$ (for $K=2$, in their case).


Although the above mentioned approximations are easy to compute, their main limitation is that they assume $\rho$ is equal for all the methods. In other words, they assume the unlikely case that each method has equal complexity and generalization capability. Indeed, in Figure \ref{fig:german_age_av} we clearly observe that different methods have different stabilities with respect to data composition. 
In order to overcome such drawback we propose an alternative complexity-aware method to estimate $\rho$, described below.

\subsubsection{Approximating the correlation $\rho$}
Consider a reference algorithm $\mathcal{M}_0$ in the particular setting under consideration, with correlation $\rho_0$ (e.g. SVM, whose $\rho_0$ can be approximated accurately by the proposal of \citet{nadeau2003inference} as detailed in the previous section). 
Then, the estimated correlation $\rho_1$ of any algorithm $\mathcal{M}_1$ under a $K$-fold CV framework is described with respect to the reference $\rho_0$ by:

\begin{equation}
    \rho_1 = \frac{(r-1)+r(K-1)\rho_0}{K-1},
    \label{eq:rho_rel}
\end{equation}
where $r = \frac{\textrm{Var}[\Bar{\mu}_{K,\mathcal{M}_0}]}{\textrm{Var}[\Bar{\mu}_{K,\mathcal{M}_1}]}$ refers to the ratio between the variances of both methods on a given metric $\mu$ in a $K$-fold CV. We note that those variances cannot be directly calculated and need to be approximated. Nonetheless, \citet{nadeau2003inference} propose an ultra conservative overestimation of the variance, from which the ratio $r_{over}=\frac{\sigma_{over, \mu, K, \mathcal{M}_0}^2}{\sigma_{over, \mu, K, \mathcal{M}_1}^2}$ can be concluded. We refer to Appendix \ref{sec:rho_explanation} for a detailed explanation of equation \eqref{eq:rho_rel} and the variance overestimation procedure. Here, we assume such an overestimation is proportional for all the methods, that is, $  r = \frac{\sigma_{\mu, K, \mathcal{M}_0}^2}{\sigma_{\mu, K, \mathcal{M}_1}^2} \approx r_{over} .$
Furthermore, note that this approximation is given in terms of the reference correlation $\rho_0$, for which generally a good approximation or a range of possible values is available as mentioned above (typically $\rho_0=1/K$ or $\rho_0 \in [0, 1/K]$). Thus, Eq. \eqref{eq:rho_rel} provides not only a pointwise estimation of $\rho$, but also an upper bound.

\section{Fairness and Uncertainty-Aware Comparison Between Two Methods}
\label{sec:comparison}


In this section we propose a new criterion for the comparison of two different algorithms, say $\mathcal{A}$ and $\mathcal{B}$, in terms of their performance and fairness guarantees, described through a set of metrics $\{\boldsymbol\Theta, \boldsymbol\eta\}$. Consider $\Delta\boldsymbol\Theta = \boldsymbol\Theta_\mathcal{A} - \boldsymbol\Theta_\mathcal{B}$ and $\Delta\boldsymbol\eta = \boldsymbol\eta_\mathcal{A} - \boldsymbol\eta_\mathcal{B}$ the performance and fairness gap, respectively, between the two algorithms, and denote by $\delta \sim P(\Delta\boldsymbol\Theta, \Delta\boldsymbol\eta)$ their joint distribution. We generalize the notion of Region of Practical Equivalence (RoPE; \citet{benavoli2017time}) as the volume around the origin that will represent the values of $(\Delta\boldsymbol\Theta, \Delta\boldsymbol\eta)$ that are considered negligibly indifferent from $\mathbf{0}\in \mathbb{R}^{N+M}$. 

 


\begin{definition} Given $\boldsymbol\epsilon_{\Theta} \in \mathbb{R_+}^N$ and $\boldsymbol\epsilon_{\eta} \in \mathbb{R_+}^M$, the Region of Practical Equivalence (RoPE) of size $(\boldsymbol\epsilon_{\Theta}, \boldsymbol\epsilon_{\eta})$ is
\begin{align*}
    RoPE(\boldsymbol\epsilon_{\Theta} ,\boldsymbol\epsilon_{\eta} )=\{(\mathbf{x},\mathbf{y}) \in \mathbb{R}^{N+M}:
   |x_i | \leq {\epsilon_{\Theta}}_i , |y_j | \leq {\epsilon_{\eta}}_j  \}.
\end{align*}
\end{definition}

\begin{figure}[ht]
\begin{center}
\centerline{\includegraphics[width=0.5\textwidth]{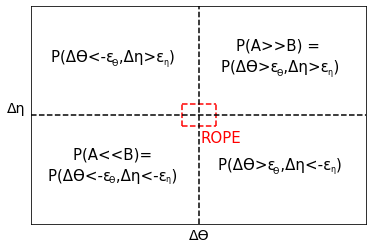}}
\caption{Graphical representation of a UM-based comparison between algorithms $\mathcal{A}$ and $\mathcal{B}$, grounding on a set of 2 metrics $\{ \Theta, \eta \}$, e.g. to measure performance ($\Theta$) and fairness guarantees ($\eta$).}
\label{fig:example_rope}
\end{center}
\end{figure}

The size of the RoPE $(\boldsymbol\epsilon_{\Theta}, \boldsymbol\epsilon_{\eta})$ will depend on the application domain.
Then, the comparison between $\mathcal{A}$ and $\mathcal{B}$ is based in 
the relative position between $(\Delta\boldsymbol\Theta, \Delta\boldsymbol\eta)$ and the RoPE (see Figure \ref{fig:example_rope}). Mainly, we are interested in estimating the posterior probabilities that:
\begin{itemize}
    \item[(a)]  $\mathcal{A}$ and $\mathcal{B}$ are practically equivalent: $$P(\mathcal{A}\approx \mathcal{B})=P((\Delta\boldsymbol\Theta, \Delta\boldsymbol\eta) \in RoPE(\boldsymbol\epsilon_{\Theta} ,\boldsymbol\epsilon_{\eta} )),$$
as the proportion of the volume of $\delta$ that intersects with the RoPE; 
 \item[(b)] $\mathcal{A}$ practically outperforms $\mathcal{B}$ in all objectives: 
$$P(\mathcal{A} >> \mathcal{B}) = P\left((\Delta\boldsymbol\Theta, \Delta\boldsymbol\eta) \subset \mathbb{R_+}^{N+M} \cap RoPE^c(\boldsymbol\epsilon_{\Theta} ,\boldsymbol\epsilon_{\eta} ) \right),$$
as the proportion of the volume of $\delta$ that intersects with the strictly positive cone that is outside the RoPE.
 
  \item[(c)] $\mathcal{B}$ practically outperforms $\mathcal{A}$ in all objectives: 
$$P(\mathcal{A} << \mathcal{B}) = P\left((\Delta\boldsymbol\Theta, \Delta\boldsymbol\eta) \subset \mathbb{R_-}^{N+M} \cap RoPE^c(\boldsymbol\epsilon_{\Theta} ,\boldsymbol\epsilon_{\eta} ) \right),$$
as the proportion of the volume of $\delta$ that intersects with the strictly negative cone that is outside the RoPE.
\end{itemize}

We look for high values in such probabilities ensuring those remarkable events. Otherwise, the probabilities of all the rest of possible events (such as $\mathcal{A}$, resp. $\mathcal{B}$, outperforming $\mathcal{B}$, resp. $\mathcal{A}$, only in a subset of the objectives, while being outperformed by $\mathcal{B}$, resp. $\mathcal{A}$, in the others) could be estimated analogously as the volume of $\delta$ intersecting the corresponding cone. 

\section{Related Work}
\label{sec:related}

We describe related work in two areas: the existing ways to assess fairness results and Bayesian methods in the context of fair ML.


\textbf{On the assessment of fairness results}.
Despite the vast amount of work that has been developed in the field of algorithmic fairness, there is no consensus on the optimal assessment of fairness results.
 In most cases, the fairness and accuracy of various models are reported \emph{separately} based on average values of repeated experiments \citep{QiaPhaLutHuetal21}, but the information provided is rather limited and oftentimes unstable \cite{friedler2019comparative}. 
 Furthermore, experimental results are shown for specific choices of accuracy and fairness trade-offs that are not particularly well justified. 
 A fair sample from the Pareto frontier is really what is needed to be able to confidently make statements about how different approaches compare.
\citet{agarwal2018reductions} produced the convex envelope of the classifiers obtained on training data at various accuracy–fairness trade-offs. 
 \citet{ji2020can} propose to use unlabeled data to get better estimates of fairness metrics, whereas \citet{romano2020achieving} make use of randomization tests. Some of the assessment schemes are context dependent and are only suitable to draw conclusions in those particular scenarios. Instead, we propose to assess the results in a context-independent manner by means of posterior distributions, similar to the work by \citet{ji2020can}, but further extending their approach in two ways: on the one hand, with the possibility of defining a $N$-dimensional joint posterior distribution of any combination of performance and fairness metrics, and not limiting the analysis to marginal distributions; and on the other hand, by allowing more complex frameworks to evaluate learning algorithms, such as $K$-fold CV, where $K$ correlated results are obtained.

\textbf{On Bayesian methods in fairness}.
The incorporation of uncertainty in the fairness-aware context has mainly been concentrated on model uncertainty \citep{dimitrakakis2019bayesian, FouIslKeyPan20}, considering a probabilistic distribution over model parameters. This concern has most popularly been addressed by means of Bayesian Neural Networks (BNNs) \citep{bhatt2021uncertainty} recently, treating model weights as random variables whose probability distributions are updated by means of Bayesian inference.
 Several ensemble-based approaches have also been proposed \citep{FouIslKeyPan20}. The uncertainty modeled in those cases is related to the confidence of the algorithms on their predictions. However, our work is different in that it instead models the uncertainty inherent in the performance and fairness metrics used to report the behavior of the algorithms. 

\section{Experiments}
\label{sec:experiments}

In this section, we present several numerical experiments carried out to validate UM and highlight its main advantages over existing procedures for algorithmic comparison in a bias-aware context. On the one hand, we analyze several case studies of hold-out evaluations in which our method not only provides further and more insightful information than the conventional results, but also provides different conclusions than those drawn under such evaluation framework. On the other hand, with respect to the $K$-fold CV evaluation framework, we first study which is the best approximation of the $\rho$ in \eqref{eq:effective_k} and, therefore, for the effective CM, on several benchmark datasets and fairness-enhancing interventions. Eventually, we consider multiple case studies in which the UM framework applied to the 10-fold CV procedure provides more complete and stable information than the classical setting.

The methods and datasets selected for this section are inspired by \citet{donini2018empirical}, since it collects diverse results w.r.t. different benchmark datasets and methods and further, they consider both hold-out and 10-fold CV scenarios. We have replicated their hold-out and $10$-fold CV results; although the results are not identical, they are very close. A complete description of benchmark algorithms and datasets that are used for the empirical evaluation can be found in the Appendix \ref{sec:ap_exp}, along with further implementation details. Moreover, in all the experiments, we have considered the uniform Dirichlet prior $\textrm{Dir}(1,1,1,1)$ for the multinomial parameter $\pi_s$.

\subsection{Hold-out}
\label{sec:exp_ho}

This section shows how the conclusions drawn from a hold-out framework vary when performance metrics are assumed to be random variables.
The experiments are performed with the Adult Income dataset which has a given train/test partition. Besides, we consider the set of method of the experiments from \citet{donini2018empirical}: SVM as a baseline method with both linear and non-linear kernel, the post-processing method by \citet{hardt2016equality} (denoted as HardtPP) combined with both baseline methods, the linear and non-linear versions of the Fair Empirical Risk Minimization (FERM) by \citet{donini2018empirical} and the in-processing method by \citet{zafar2017fairness}. The algorithms are compared in terms of accuracy $\Theta$ and Equality of Opportunity (EOp) $\eta=TPR_1-TPR_0$. Furthermore, we have chosen a square ROPE with dimensions $0.01 \times 0.01$.  
\begin{figure*}[ht!]
\includegraphics[width=\textwidth]{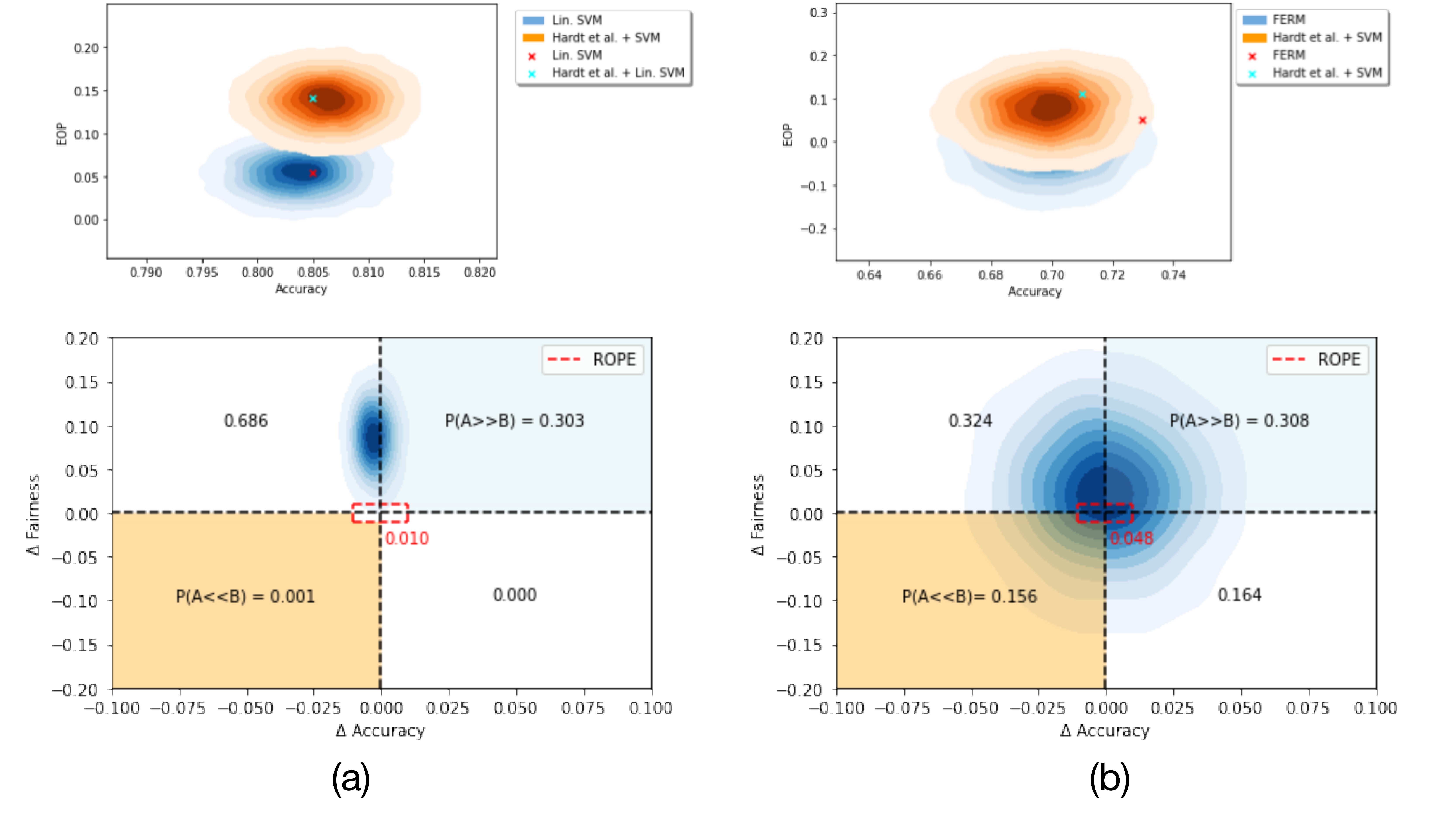}
\caption{\textbf{(a)}: Hold-out results on Adult Income dataset. 
\textbf{(b)}: K-fold CV results on German dataset. 
\textbf{(Top)}: The joint posterior distribution of accuracy and EOp for two methods with crosses show the hold-out/K-fold CV results. 
\textbf{(Bottom)}: The posterior distribution of the performance gap $\delta$ between the performances of the algorithm.
In all plots, the x-axis denotes accuracy (re. the difference in accuracy) and the y-axis reflects fairness guarantees (re. the difference in fairness), in this case, measured by the EOp fairness metric. 
In \textbf{(a)}, the two compared methods are: \textbf{(a-1)}: SVM with linear kernel (blue) and \textbf{(a-2)}: the post-processing fairness method  HardtPP
In \textbf{(b)}, the two methods are: \textbf{(b-1)} the in-processing fairness method FERM (blue) and \textbf{(b-2)} the post-processing fairness method HardtPP.
Although classical results based on hold-out/K-fold CV suggest that \textbf{(a-1)}/\textbf{(b-1)} outperforms \textbf{(a-2)}/\textbf{(b-2)}, according to our uncertainty matters UM framework, the probability for outperforming is only 0.3./0.305, respectively.\label{fig:case1_case2}}
\end{figure*}

\textbf{Case study 1: ($\mathcal{A}$) Linear SVM and ($\mathcal{B}$) Non-linear SVM + HardtPP} According to the classic result, both methods have an accuracy of 0.805, but ($\mathcal{A}$) provides better fairness guarantees than ($\mathcal{B}$): method ($\mathcal{A}$) obtains an EOp value of 0.05 while ($\mathcal{B}$) goes up to 0.14. With that, ($\mathcal{A}$) would be preferred to ($\mathcal{B}$), since for equivalent accuracy it provides better fairness guarantees. On the other hand, under the UM framework, from which we obtain the posteriors and the distribution of the difference shown in Figure \ref{fig:case1_case2}(a), we conclude that the probability of ($\mathcal{A}$) outperforming ($\mathcal{B}$) in accuracy and EOp is only $P(\mathcal{A}>>\mathcal{B}) = 0.30$. Most of the area covered by $\delta$, is concentrated on the second quadrant, from which we can estimate that there is a probability of 0.69 that method ($\mathcal{B}$) has better accuracy but is outperformed by method ($\mathcal{A}$) in terms of fairness.

\textbf{Case study 2: ($\mathcal{A}$) Non-linear SVM and ($\mathcal{B}$) FERM.} Under the classic procedure, method ($\mathcal{A}$) obtains an accuracy of 0.84 and its EOP is of 0.07; while method ($\mathcal{B}$) shows a predictive accuracy of 0.83 and the value of EOp is 0.09. Based on this results, method ($\mathcal{A}$) outperforms ($\mathcal{B}$) in both objectives. However, the analysis based on UM suggests more fine-grained conclusions: the probability of such dominance is only of 0.41. Besides, the probability of equivalence is of $P(\mathcal{A} \approx \mathcal{B}) = 0.2$, and so is the probability that ($\mathcal{A}$) has better predictive performance but worse fairness guarantees. 

Additional case studies and graphical results can be found in Appendix \ref{sec:ap_ho_res}.

\subsection{K-fold CV}

In this section, we evaluate the UM framework in the $K$-fold CV evaluation setting.  We first must identify the optimal strategy for estimating $\rho$ to apply it afterwards in several case studies of algorithmic comparison as in the previous section. For the experiments of this section we use the German Credit dataset, for which \citet{donini2018empirical} provide classical 10-fold CV results in terms of accuracy ($\Theta$) and EOp ($\eta$=EOp=$TPR_1-TPR_0$).

\subsubsection{Best approximation for $\rho$}
\label{sec:exp_rho}

\begin{table*}[ht!]
\caption{ The area of the 95\% HDR of  $\delta$ and the proportion of 10,000 different 10-fold CV results it encloses (\%res) for different algorithms, for approximations (1)-(4) to estimate the correlation $\rho$. We use the German Credit dataset with sex as the sensitive attribute. }
\label{tab:german_sex_eop}
\begin{center}
\begin{small}
\begin{sc}
\begin{tabular}{lcccccccc}
\toprule
&  \multicolumn{2}{c}{$\rho=1/K$} & \multicolumn{2}{c}{$\rho\in[0,1/K]$} & \multicolumn{2}{c}{$\rho_{rel}$} & \multicolumn{2}{c}{$\rho_{rel}^{\uparrow}$} \\
Method & Area & \% res & Area & \% res & Area &  \% res & Area &  \% res   \\
\midrule
LR & 0.0207 & 100.0 & 0.0153 & 100.0 & 0.0197 & 100.0 & 0.0150 & 99.96   \\
SVM & 0.0201 & 100.0 & 0.0148 & 99.99 & 0.0201 & 100.0 & 0.0148 & 99.99  \\
Linear SVM & 0.0205 & 99.98 &  0.0154 & 99.86 & 0.0243 & 100.0 & 0.0168  &  99.87  \\
\citet{kamiran2012data} + LR & 0.0207 & 100.0 & 0.0153 & 100.0 & 0.0195 & 100.0 & 0.0148 & 100.0 \\ 
\citet{kamiran2012data} + SVM & 0.0201 & 100.0 & 0.0149 & 99.96 & 0.0196 & 99.95 & 0.0148 & 99.64 \\
\citet{feldman2015certifying} + LR & 0.0208 & 100.0 & 0.0154 & 100.0 & 0.0182 & 99.99 & 0.0143 & 99.91  \\
\citet{feldman2015certifying} + SVM & 0.0201 & 100.0 & 0.0149 & 99.99 & 0.0198 & 99.99 & 0.0147 & 99.93  \\
\citet{zafar2017fairness} & 0.0205 &  100.0 & 0.0153  & 100.0 & 0.0269 & 100.0 & 0.0178 & 99.99  \\
FERM \cite{donini2018empirical} & 0.0107 & 100.0 & 0.0079 & 99.74 & 0.0110 & 100.0 & 0.0080 & 99.86  \\
LFERM \cite{donini2018empirical} & 0.0093 & 99.99 & 0.0069 & 99.93 & 0.0112 & 100.0 & 0.0075 & 99.97   \\
LR + \citet{hardt2016equality} & 0.0210 & 90.32 & 0.0156 & 82.08 & 0.0414 & 99.04 & 0.0232 & 92.77 \\
SVM + \citet{hardt2016equality} & 0.0203 & 94.09 & 0.0151 & 88.67 & 0.0413 & 99.48 & 0.0230 & 95.84 \\
Linear SVM + \citet{hardt2016equality} & 0.0196 & 92.91 & 0.0145 & 86.94 & 0.0904 & 99.99 & 0.0381 & 99.00 \\
\bottomrule
\end{tabular}
\end{sc}
\end{small}
\end{center}
\end{table*}

First, we compare the existing alternatives to approximate $\rho$ to find the most accurate strategy. The most precise approximation for $\rho$ will be the one that provides the narrowest posteriors with high degree of confidence. For that purpose, starting from an arbitrary initial 10-fold CV we compute the posterior distributions obtained with the different approximations. For each posterior distribution, we estimate the area of the 95\% \textit{highest density region} (HDR) \cite{hyndman1996computing} and the proportion of repeated 10-fold CV results that fall within such region (\% res). The former is a very compact summary of the most credible values of a random variable (additional information can be found in the Appendix \ref{sec:ap_exp}). The narrowest posterior able to enclose all possible 10-fold CV results will be the optimal strategy for approximating $\rho$. We repeat the experiment for 10,000 different 10-fold CV configurations, reporting results averaged over all of them. A more detailed characterization of the experimental setup can be found in Appendix \ref{sec:ap_exp}. Besides, regarding the algorithms considered, apart from those listed in Section \ref{sec:exp_ho}, we have also considered Logistic Regression (LR) as a baseline classifier, and also, an additional pre-processing strategy proposed by \citet{feldman2015certifying}.

We have considered several alternatives to approximate $\rho$: (1) $\rho=1/K$; (2) $\rho \in [0, 1/K]$, (3) the relative $\rho$ assuming that $\rho_0=1/K$ and $\mathcal{M}_0=$ SVM (non-linear) (denoted as $\rho_{rel}$); and (4) the relative $\rho$ assuming that $\rho_0 \in [0, 1/K]$ and $\mathcal{M}_0=$ SVM (non-linear) (denoted as $\rho_{rel}^{\uparrow}$). Table \ref{tab:german_sex_eop} summarizes the results: on the one hand, in the cases where methods (1) and (2) estimate the uncertainty well, the relative procedure optimizes the uncertainty estimation, providing slightly more precise estimations (e.g. in the cases of the pre-processing methods by \citet{kamiran2012data} and \citet{feldman2015certifying}, and the in-processing method by \citet{donini2018empirical}). However, the significant improvement can be seen in the case of the post-processing method by \citet{hardt2016equality}, where the uncertainty is not estimated well by means of conventional approximations (1) and (2) (the number of real 10-fold CV results that is covered by the HDR region of the posterior distribution falls down to 82\% for some cases), while our alternative achieves high confidence estimations. Therefore, the alternatives newly proposed in the paper (i.e, (3) and (4)) provide more precise uncertainty estimations with equivalent degree of confidence, outperforming conventional approximations for $\rho$. Besides, the high degree of confidence highlights that with a single 10-fold CV we are able to cover all the results corresponding to each possible 10-fold splitting configuration (recall Figure \ref{fig:german_age_av}) with very high probability. 

Moreover, the uncertainty estimations agree with the reported uncertainty estimations in the 10-fold CV in \citet{donini2018empirical}: the higher standard deviation shown by the results of a method, the higher the uncertainty according to the statistical framework (wider HDR). Indeed, the post-processing method by \citet{hardt2016equality} has the highest uncertainty according to our framework, as well as the highest standard deviation in the results from \cite{donini2018empirical}, and the biggest disparity in the results in Figure \ref{fig:german_age_av}. Also, the standard deviation is the smallest in the case of the linear and non-linear versions of FERM, and so are the areas of the HDR\textquotesingle s of their respective posterior distributions. However, note that the standard deviation of the $K$-fold tends to underestimate the uncertainty of the results, a limitation that can be solved by the UM framework.

\subsubsection{Classic vs. UM Conclusions}
\label{sec:exp_cv}

This section highlights the difference between the conclusions drawn under the classic and UM evaluation frameworks in the case of the 10-fold CV. We have considered the same set of methods from \ref{sec:exp_ho}. Furthermore, the posterior distributions will be calculated using $\rho_{rel}^{\uparrow}$ as the approximation for the correlation with $\rho_0 \in [0, 1/K]$ and $\mathcal{M}_0=$ SVM (non-linear), which has shown to provide the most precise uncertainty estimations. 


\textbf{Case study 1: ($\mathcal{A}$) FERM and ($\mathcal{B}$) Non-linear SVM + HardtPP.} According to the reported results, the method ($\mathcal{A}$) seemingly outperforms ($\mathcal{B}$): it has better predictive accuracy (0.73 vs. 0.71) and better fairness guarantees (0.05 vs. 0.11). However, under the UM result assessment, the probability of such event is only 0.305. Besides, there is a probability of 0.157 that ($\mathcal{B}$) outperforms ($\mathcal{A}$) in all the objectives. Furthermore, the probability that ($\mathcal{A}$) has better accuracy but worse fairness guarantees is equal to the probability of the opposite situation (i.e. ($\mathcal{B}$) has higher accuracy but worse fairness guarantees), which is 0.325. 

\textbf{Case study 2: ($\mathcal{A}$) Linear FERM and ($\mathcal{B}$) Linear SVM + HardtPP.} Classical results suggest the linear version of the FERM method also significantly outperforms the combination between the SVM with linear kernel and the post-processing method HardtPP: the predictive performance and the fairness guarantees are considerably better (0.69 vs. 0.61 in accuracy and 0.05 vs. 0.15 in EOP). Nonetheless, according to the framework, the probability $P(\mathcal{A}>>\mathcal{B})$ is 0.23. What is more, the probability of outperforming $P(\mathcal{B}>>\mathcal{A})$ is also similar: 0.22. Therefore, there is no clear evidence of dominance towards any of the algorithms. It is worth noting that, in this case study, all the possible outcomes except $\mathcal{A} \approx \mathcal{B}$ have the same probability of around 0.23.

\textbf{Case study 3: ($\mathcal{A}$) Non-linear SVM + HardtPP and ($\mathcal{B}$) Linear SVM + HardtPP.} From the classical empirical results, it could be stated that using a non-linear kernel for SVM with the post-processing method HardtPP provides not only more accurate predictions (0.71 vs. 0.61) but also fairer outcomes in terms of EOP (0.11 vs 0.15); that is, ($\mathcal{A}$) outperforms ($\mathcal{B}$). Nonetheless, the UM analysis provides different conclusions: $P(\mathcal{A}>>\mathcal{B}) = 0.144$ while $P(\mathcal{B}>>\mathcal{A})=0.322$. Therefore, there is a higher probability that ($\mathcal{B}$) outperforms ($\mathcal{A}$) than the opposite case of ($\mathcal{A}$) outperforming ($\mathcal{B}$).

Additional graphical results can be found in Appendix \ref{sec:ap_cv_res}. 

\section{Conclusions and Future Works}
\label{sec:conclusion}

We have proposed a probabilistic approach based on Bayesian inference for the uncertainty-aware assessment of fairness results w.r.t. multiple objectives, called UM, which can be applied under different evaluation settings (e.g, hold-out and $K$-fold CV). As a valuable application of our method, we are able to estimate the uncertainty-aware performance and fairness gap existing between two algorithms. The numerical experiments have shown that the UM framework is more informative and widens the perspective regarding the behavior of the algorithms. Besides, they highlight that UM captures the unstable nature of the results, and provides an alternative assessment allowing to derive stable conclusions. Whenever the probabilistic comparison does not show high probability regarding any of the possible outcomes, it is an indication that more data needs to be incorporated to the evaluation process to derive statements in favor of any of those possibilities.  

One of the main drawbacks of our method to approximate the correlation between the results of the $K$-fold CV is that, for very large datasets, it might result too time-consuming to be applicable under given settings. Nonetheless, the conventional approximation becomes accurate for increasing dataset size and can be employed as an alternative. Still, our approximation is the most accurate in the low-data regime, where the quantification of uncertainty becomes more relevant. It is also worth mentioning that, although the choice of the RoPE may slightly affect the probability of the events, it does not have an effect on the main conclusion of algorithmic comparison. Besides, for a given difference distribution, the users can apply the RoPE they consider appropriate to draw their own conclusions. 

We acknowledge that the uncertainty aware assessment of performance and fairness guarantees would be beneficial in the development of novel fairness-enhancing interventions. Moreover, UM could be extended to further fairness-aware evaluation frameworks by, e.g., deriving uncertainty-aware pareto frontiers.

\addcontentsline{toc}{section}{References}

\printbibliography

\newpage

\appendix

\section{Unstable 10-fold CV Results of Fairness-Enhancing Interventions.}

\begin{figure}[h!]
    \centering
    \includegraphics[width=\textwidth]{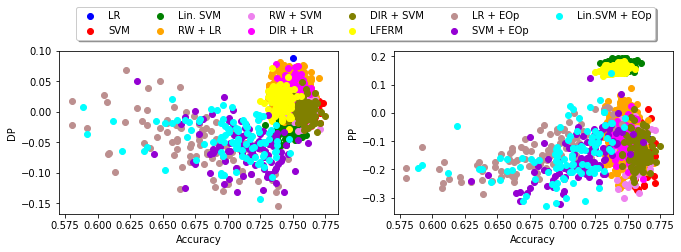}
    \caption{10-fold CV results regarding accuracy and fairness, measured by means of (left) Demographic Parity (DP) and (right) Predictive Parity (PP), for 10,000 different 10-fold CV configurations. Each point refers to one 10-fold CV result, and the colors refer to different benchmark fairness-enhancing methods. These results were obtained using the German dataset with age as the sensitive attribute.}
    \label{fig:german_age_av_2}
\end{figure}

\section{Proofs of Section \ref{sec:uncertainty}}
\label{sec:proofs}

 In order to proof Proposition \ref{prop:effectivecm} we first introduce the reader to an important lemma by \citet{nadeau2003inference}:
    \begin{lemma}
        Let $U^1$, ... $U^K$ be random variables with common mean $\beta$ and the following covariance structure
        \begin{equation*}
            \textrm{Var}[U^k] = \delta \quad \forall k, \quad \textrm{Cov}[U^k, U^{k'}] = \gamma \quad \forall k \neq k'.
        \end{equation*}
        Let $\pi = \frac{\gamma}{\delta}$ be the correlation between $U^k$ and $U^{k'}$ ($k \neq k'$). Let $\Bar{U} = k^{-1} \sum_{k=1}^K U^k$ and $S_{U}^2 = \frac{1}{K-1} \sum_{k=1}^K (U^k - \Bar{U})^2$ be the sample mean and the sample variance respectively. Then:
        \begin{enumerate}
            \item $\textrm{Var}[\Bar{U}] = \gamma + \frac{(\delta-\gamma)}{K} = \delta \big( \pi + \frac{1-\pi}{K} \big)$.
            \item If the stated covariance structure holds for all $K$ (with $\gamma$ and $\delta$ not depending on $K$), then:
            \begin{itemize}
                \item $\gamma \geq 0$
                \item $\lim_{K \rightarrow \infty} \textrm{Var}[\Bar{U}] = 0 \iff \gamma = 0$
            \end{itemize}
            \item E$[S_{U}^2] = \delta - \gamma$
        \end{enumerate}
        \label{lemma:1}
    \end{lemma}
    
\textbf{Proof of Proposition \ref{prop:effectivecm}}
\begin{proof}
    Although we are omitting $s$ for clarity, note that this holds for any $CM$ or $CM_s$.

    Let $\mu$ be a metric of interest (e.g. accuracy, TPR, equality of opportunity), $\hat{\mu}$ its mean value resulting from the $K$-fold cross-validation and $\mu^k$ its value for the $k$-th train/test configuration of the cross-validation. We will assume that the variance of the result of each train/test configuration of the $K$-fold cross-validation will be similar, i.e. $\textrm{Var}[\mu^k]$ similar across all $k$. Furthermore, we will assume that the covariance between the results of two different train/test configurations on the $K$-fold cross-validation will be similar for any two pairs, i.e.  $\textrm{Cov}[\mu^k, \mu^{k'}] =\rho$ for $k \neq k'$. Then, from Lemma \ref{lemma:1} we know that:
    \begin{equation*}
        \textrm{Var}[\hat{\mu}] = \frac{\textrm{Var}[\mu^k] (1+(K-1)\rho)}{K}
    \end{equation*}

    Let us consider the case where $\mu$ is TPR. Then:
    \begin{equation*}
        \textrm{Var}[\mu^k] = \textrm{Var}[TPR^K] =  \textrm{Var} \bigg[ \frac{TP^k}{TP^k + FN^k}  \bigg]
    \end{equation*}
    $TP^k + FN^k$ is constant, since it must add up to the number of positive labeled instances in the test set. In the case of a $K$-fold cross-validation that would be $n_{+}/K$ where $n_{+}$ refers to the number of positive labeled instances in the whole dataset. Thus,
    \begin{equation*}
        \textrm{Var}[TPR^K] = \bigg ( \frac{K}{n_{+}} \bigg )^2  \textrm{Var} [TP^k]
    \end{equation*}
    The marginals of the Multinomial distribution are binomial, thus the counts follow binomial distributions. Besides, from \citet{goutte2005probabilistic} we know that for fixed $TP^k + FN^k$, $TP^k$ follows a binomial distribution with parameters $r$ and $n_{+}/K$, that is, $TP^k | TP^k + FN^k \sim \textrm{Bin}(r, n_{+}/K)$. Therefore, $\textrm{Var} [TP^k] = r(r-1)\frac{n_{+}}{10}$ and:
    \begin{equation*}
        \textrm{Var}[TPR^K] = \frac{Kr(1-r)}{n_{+}} 
    \end{equation*}
    where $r = p(\hat{y} = 1| y = 1)$. With that, the variance of the sample mean (the estimation resulting from the $K$-fold cross-validation) would be:
    \begin{equation}
        \textrm{Var}[\hat{TPR}] = \frac{r(1-r)(1+(K-1)\rho)}{n_{+}}
        \label{eq:var_cv}
    \end{equation}
    At the same time, if we describe the results obtained from a $K$-fold by means of an effective CM (composed of effective counts):
    \begin{equation*}
        \textrm{Var}[\hat{TPR}] = \textrm{Var} \bigg[ \frac{TP^e}{TP^e + FN^e}  \bigg]
    \end{equation*}
    As previously stated, $TP^e + FN^e$ is constant and $TP^e | TP^e + FN^e \sim \textrm{Bin}(r, TP^e + FN^e)$. Thus,
    \begin{equation}
        \textrm{Var}[\hat{TPR}] = \frac{r(1-r)}{TP^e + FN^e}
        \label{eq:eff_var}
    \end{equation}
    Comparing equations (\ref{eq:var_cv}) and (\ref{eq:eff_var}) we get:
    \begin{equation*}
        \frac{r(1-r)}{TP^e + FN^e} = \frac{r(1-r)(1+(K-1)\rho)}{n_{+}}
    \end{equation*}
    and since $n_{+} = \sum_{k=1}^K (TP^k + FN^k)$:
    \begin{equation*}
        TP^e + FN^e = \frac{1}{1+(K-1)\rho} \sum_{k=1}^K (TP^k + FN^k)
    \end{equation*}
    Consequently:
    \begin{equation*}
        TP^e = \frac{1}{1+(K-1)\rho} \sum_{k=1}^K TP^k
    \end{equation*}
    \begin{equation*}
        FN^e = \frac{1}{1+(K-1)\rho} \sum_{k=1}^K FN^k
    \end{equation*}
    If we repeat the procedure with the metric $FPR$ and assuming that the value of the correlation $\rho$ is equal for all the metrics, we would derive equivalent formulations for the effective counts on FP and TN and, finally obtain \eqref{eq:effective_k}.
    
\end{proof}

\section{The Relative Value of the Correlation $\rho$}
\label{sec:rho_explanation}

In this section we provide a more detailed explanation about how eq (\ref{eq:rho_rel}) is concluded. Moreover, we explain the method proposed by \citet{nadeau2003inference} to obtain the over-estimations of the variance which is used to compute the value of $r$ of our approximation.  

From Lemma \ref{lemma:1} we know that the variance of a metric under a $K$-fold CV can be written as:
\begin{equation*}
    \textrm{Var}[\Bar{\mu}_{K}] = \frac{\textrm{Var}[\mu^k] (1+(K-1)\rho)}{K}
\end{equation*}
Thus, for two ML methods ($\mathcal{M}_0$ and $\mathcal{M}_1$) with correlations $\rho_0$ and $\rho_1$, the ratio of their variances is:
\begin{equation*}
    r = \frac{\textrm{Var}[\mu^k_1]}{\textrm{Var}[\mu_0^k]} \cdot \frac{1+(K-1) \rho_1}{1+(K-1) \rho_0}
\end{equation*}
We assume the variance of the results obtained for each fold will have the same variance for each $k = 1,..., K$ (Lemma \ref{lemma:1}), and will be similar for different methods (which has been supported by different experiments); that is, $\textrm{Var}[\mu^k_1] \approx \textrm{Var}[\mu_0^k]$ (note that, starting from the same $\textrm{Var}[\mu^k]$, what makes the variance of the final metric to be different is the correlation). Thus:
\begin{equation*}
    r = \frac{1+(K-1) \rho_1}{1+(K-1) \rho_0}
\end{equation*}
From which we can conclude eq (\ref{eq:rho_rel}):
\begin{equation*}
    \rho_1 = \frac{(r-1)+r(K-1)\rho_0}{K-1},
\end{equation*}
The value $r = \frac{\textrm{Var}[\Bar{\mu}_{K,M0}]}{\textrm{Var}[\Bar{\mu}_{K,M1}]}$ cannot be estimated exactly. However, \citet{nadeau2003inference} propose a method to overestimate the variance $\textrm{Var}[\Bar{\mu}_{K}]$. Assuming that such an overestimation is proportional for all the methods, that is, $  r = \frac{\sigma_{\mu, K, M0}^2}{\sigma_{\mu, K, M1}^2} \approx r_{over}$, we can obtain an approximate value of $r$. With that, starting from a reference value $\rho_0$ we can obtain the value of the correlation for any method. \textit{But, how do we obtain the overestimations of the variances?}

In what follows, we explain the approach by \cite{nadeau2003inference} to overestimate the variance of a given performance metric. Particularly, they propose by obtaining independent observations of such statistic. To obtain such independent measurements, the dataset must be split into two disjoint datasets $D$ and $D^c$ (where $D \cap D^c = \emptyset$) of size $\lfloor \frac{n}{2} \rfloor$ ($|D| = |D^c| = \lfloor \frac{n}{2} \rfloor$) (being $n$ the size of the complete dataset). Let $\hat{\mu}$ and $\hat{\mu}^c$ be the values of the statistic of interest when a $K$-fold CV is performed in $D$ and $D^c$, respectively. Then, $\frac{1}{2}(\hat{\mu}-\hat{\mu}^c)^2$ is an unbiased estimate of $\sigma_{over, \mu, K}^2$. The step of splitting the dataset into two disjoint blocks can be repeated $M$ times, yielding the pairs $(\hat{\mu}_m, \hat{\mu}_m^c)$ for $m = 1,..., M$. With that, the following unbiased estimation of $\sigma_{over, \mu, K}^2$ is concluded:
\begin{equation}
    \sigma_{over, \mu, K}^2 = \frac{1}{2M} \sum_{m=1}^M (\hat{\mu}_m - \hat{\mu}_m^c)^2
\end{equation}

\begin{figure}[ht]
\vskip 0.2in
\begin{center}
\centerline{\includegraphics[width=0.4\columnwidth]{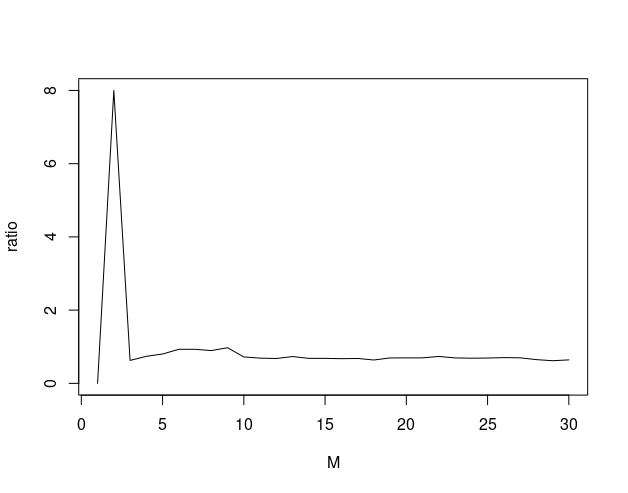}}
\caption{The ratio between the overestimation of the variances ($r_{over}$) between FERM of \citet{donini2018empirical} and non-linear SVM. The value of $r_{over}$ stabilizes for very low values for $M$.}
\label{fig:ratio_over}
\end{center}
\vskip -0.2in
\end{figure}

Thus this approximation requires performing $2M$ additional half-sized 10-fold CV procedures. Nonetheless, low values of $M$ provide stable result with respect to $r_{over}$. In fact, Figure \ref{fig:ratio_over} suggests that, in the case of the German Credit dataset (composed of 1,000 instances), $M=5$ already constitutes a good approximation. Furthermore, the value of $M$ required to obtain an accurate estimation decreases considerably for increasing dataset size. Thus, for big enough datasets $M=1$ is sufficient. Nonetheless, in such cases, the conventional approximation $\rho=1/K$ becomes accurate, not requiring the computation of any additional 10-fold CV. 

\section{Further details regarding the experiments}
\label{sec:ap_exp}

\subsection{Datasets}

In the experimental section, we consider different classification tasks that are highly popular in the literature regarding algorithmic fairness. In what follows, we describe the datasets employed in those applications. 

\textbf{Adult Income.} It is a dataset based on the data from the 1994 US Census, where the main goal is to predict whether an individual earns more than 50,000\$ per year. The 14 features used to describe the instances include occupation, marital status and education. Furthermore, it contains sensitive information such as, age, gender and race. In our experiments, we considered a single sensitive attribute: gender. This dataset is publicly available in the UCI repository\footnote{\texttt{http://archive.ics.uci.edu/ml/index.php}} and it is already divided into a training and a test set. The former has 32,561 instances and, the latter, 16,281. We pre-process both datasets as in \cite{donini2018empirical}.

\textbf{German Credit.} The German Credit dataset collects information about several individuals created from a German bank\textquotesingle s data from 1994. It contains details about the socioeconomic situation of individuals: namely its employment, housing, savings, etc. Besides, the set of features includes some sensitive information as well: the gender and age. In this classification task, the objective is to predict whether an individual should obtain a good or bad credit score. This dataset is considerably smaller than the previous, containing only 1,000 instances and 20 features, and it is publicly available in the UCI repository.

\subsection{Methods}

In this section we provide a brief description of the fairness-enhancing interventions that are considered in the experimental section. The considered approaches are grouped, according to the step in which fairness guarantees are enforced within the algorithmic procedure, into pre-processing, in-processing and post-processing methods. Pre-processing mechanisms aim to transform biased datasets so that, when conventional ML classifiers are trained on them, the final outputs are fair. In-processing interventions, modify existing algorithms to account for fairness guarantees at training time. Lastly, post-processing methods alter algorithmic predictions to obtain fairer final decisions. 

\subsubsection{Pre-processing}

\textbf{Reweighing (RW).} This pre-processing intervention proposed by \citet{kamiran2012data} aims to transform a biased dataset so that when a conventional ML classifier is fed with such data the effects of it\textquotesingle s outcomes are not disproportionate for different sensitive groups. Such transformation constitutes of weighing the instances to achieve equal prevalence across sensitive groups, i.e. to enforce statistical independence between the label and the sensitive attribute.  

\textbf{Disparate Impact Remover (DIR).} \citet{feldman2015certifying} developed a method to pre-process a biased dataset in order to remove the \textit{disparate impact} on the effects of the algorithm across different subgroups of the population when the algorithm is fed with such dataset. The processed dataset is obtained by changing the non-sensitive attributes of the dataset that could be employed to predict the sensitive information. 

\subsubsection{In-processing}

\textbf{Fair Empirical Risk Minimization (FERM).} FERM is an in-processing method developed by \citet{donini2018empirical} which constitutes a modification of the conventional Empirical Risk Minimization (ERM) method. In particular, they propose to introduce additional constraints into the optimization problem to enforce the fulfillment of fairness guarantees by the learning algorithm. These constraints request the learning algorithm to have approximately constant conditional risks with respect to the sensitive attribute. As in their work, we consider SVM as the base learning method, using either a linear kernel (linear FERM) or a non-linear kernel (FERM). We have replicated the training and evaluation procedures explained in their paper, implemented using the code provided by the authors\footnote{\texttt{https://github.com/jmikko/fair\_ERM}}.

\textbf{Avoiding Disparate Mistreatment.} \citet{zafar2017fairness} proposed an optimization problem to learn an algorithm, by minimizing a general classification loss subject to fairness constraints. The latter forced the algorithm to achieve similar $FNR$ and $FPR$ performances across the different sensitive groups (which would mean that it does not show disparate mistreatment). In order to avoid tractability issue, they reformulate the problem using a tractable proxy by defining the disparate mistreatment using the covariance between the sensitive attributes of the individuals and the signed distance between the feature vectors of misclassified instances and the classifier decision boundary. We implemented the code provided by the authors\footnote{\texttt{https://github.com/mbilalzafar/fair-classification}} with a linear decision boundary. 

\subsubsection{Post-processing}

\textbf{\citet{hardt2016equality} post-processing.} \citet{hardt2016equality} propose a fairness-enhancing intervention that modifies the outcomes of a given predictor in order to satisfy a given fairness property defined by either \textit{equality of opportunity} or \textit{equality of odds}. In the case of binary predictors, their method flips the decisions with a given probability to satisfy the fairness criteria. On the other hand, for score-based algorithms, they suggest to modify the decision boundary to improve the fairness guarantees of the algorithm: in particular, they modify the decision threshold, assuming different (possibly randomized) thresholds for the distinct sensitive groups.

\subsection{Experimental Setup to choose the optimal approximation for $\rho$}

\begin{figure}[h!]
    \centering

\begin{tikzpicture} [-, block/.style={draw, thick, minimum height=1cm, align=center} ]
   \node[inner sep=0pt] (a) at (0,0)
    {\includegraphics[width=.2\textwidth]{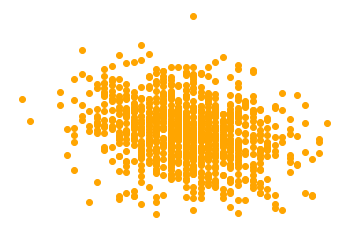}};
    \node[] at (0,1.2) (b) {10-fold CV results for different seeds for the splits};

   \node[inner sep=0pt] (c) at (-3,-2)
    {\includegraphics[width=.15\textwidth]{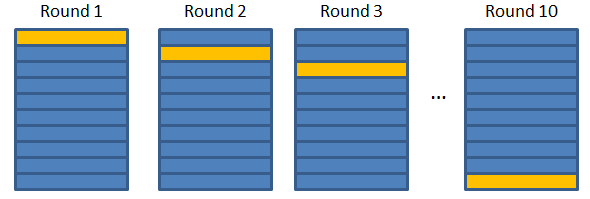}};

    \node[] at (-3, -4) (d) {$CM_s^e$};

    \node[inner sep=0pt] (e) at (0,-4)
    {\includegraphics[width=.25\textwidth]{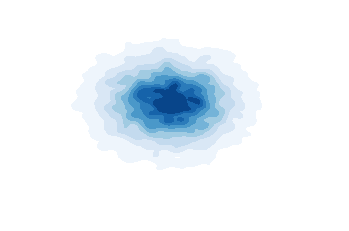}};
    
    \node[] at (-0.2,-2.8) (k) { $P(\boldsymbol\Theta, \boldsymbol\eta)$ };

    \node[inner sep=0pt] (f) at (4,-4)
    {\includegraphics[width=.25\textwidth]{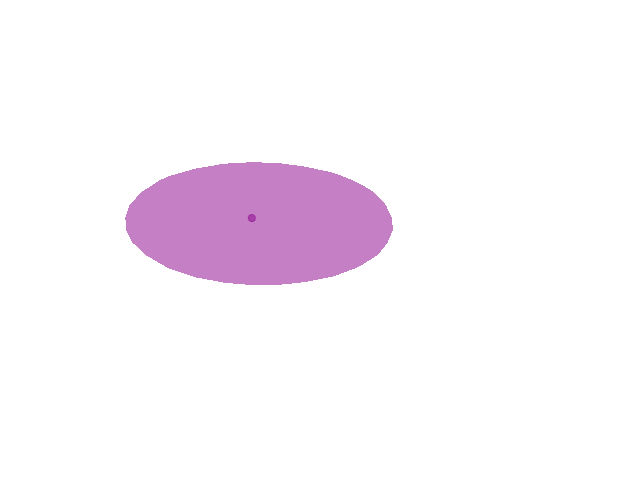}};
    
    \node[] at (3.6,-2.9) (g) {95\% HDR};

    \node[inner sep=0pt] (h) at (8,-4)
    {\includegraphics[width=.25\textwidth]{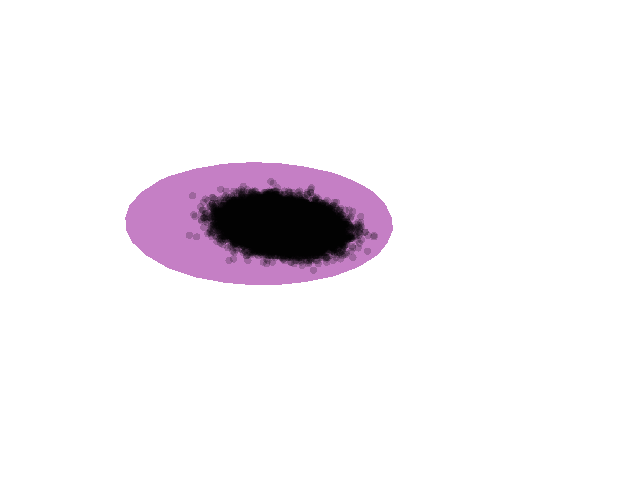}};
    \node[] at (7.6,-3.2) (i) {\% res};

     \node[] at (3.7,-5.1) (j) {Area};

     \draw [-stealth](-1.2,-0.5) -- (-3,-1.4);
     \draw [-stealth](-3,-2.7) -- (-3,-3.5);
     \draw [-stealth](-2.2,-4) -- (-1.3,-4);
     \draw [-stealth](1.3,-4) -- (2.7,-4);
     \draw [-stealth](4.8,-4) -- (6.3,-4);
     \draw [](1.5,-0.5) -- (5.5,-0.5);
     \draw [](5.5,-0.5) -- (5.5,-4);
     \draw [-stealth](3.7,-4.4) -- (3.7,-4.8);

\end{tikzpicture}

    \caption{ The workflow representing the experimental setup used to choose the best approximation for $\rho$. Starting from an arbitrary 10-fold CV configuration and compute the effective CM. From the effective CM we conclude the joint posterior distribution of accuracy and a fairness metric. Then, we estimate the 95\% HDR of the posterior distribution and calculate its area and proportion of repeated 10-fold CV results that fall within such region (\% res). We repeat this procedure for different initial 10-fold CV configurations. }
    \label{fig:setup}
\end{figure}
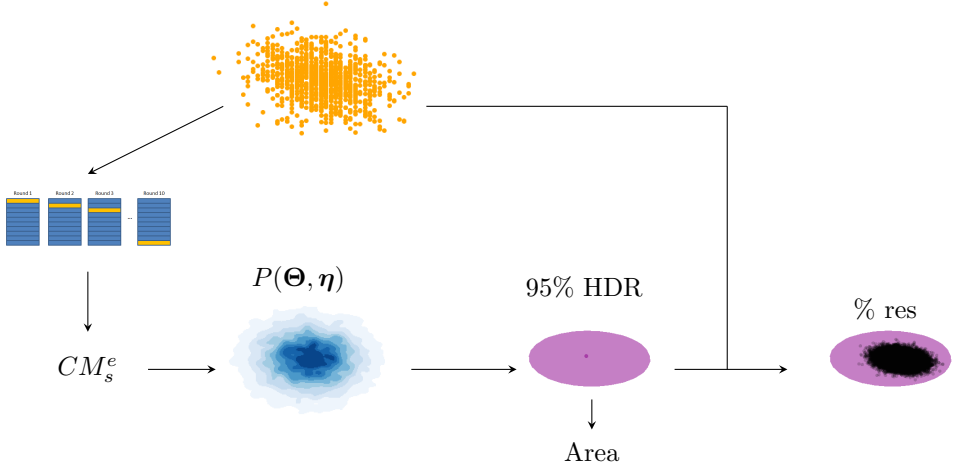

This section provides a more detailed overview regarding the experimental setup from Section \ref{sec:exp_rho}. The main goal of this experiment is to find the best approximation for the correlation $\rho$. For that purpose, starting from an arbitrary initial 10-fold CV we compute the posterior distributions obtained with the different approximations. For each posterior distribution, we estimate the area of the 95\% HDR and the proportion of repeated 10-fold CV results that fall within such region (\% res). The narrowest posterior able to enclose all possible 10-fold CV results will be the optimal strategy for approximating $\rho$. Figure \ref{fig:setup} shows a schematized version of the experimental workflow. We repeat the experiment for 10,000 different 10-fold CV configurations, reporting results averaged over all of them.

\subsection{Calculating the 95\% HDR}

In this section, we provide a formal definition of the  $(100-\alpha)\%$ HDR and outline existing alternatives to measure it in practice. 

The $(100-\alpha)\%$ \textit{highest density region} (HDR) is the $(100-\alpha)\%$ density region with the smallest size, and it constitutes the most compact summary of a probability distribution. Let $f(x)$ be the probability density function of a (possibly multivariate) random variable $X \in \mathbb{R}^d$ and $0< \alpha < 100$. Following the definition by \citet{hyndman1996computing}, the $(100-\alpha)\%$ HDR refers to the subset $\{ x \; : \; f(x) \geq f_{\alpha} \}$, where $f_{\alpha}$ is the highest constant for which $P(X \in \{ x \; : \; f(x) \geq f_{\alpha} \} ) = (100-\alpha)\%$:
\begin{equation*}
    P(X \in \{ x \; : \; f(x) \geq f_{\alpha} \} ) = \int_{ \{ x \; : \; f(x) \geq f_{\alpha} \} } f(u) \; du \; = \; (100-\alpha)\%
\end{equation*}

There are many alternatives to estimate the $(100-\alpha)\%$ HDR of a distribution. If the probability distribution $f(x)$ is known, the problem can be solved by numerical integration as in \citet{hyndman1996computing}. However, this procedure becomes computationally hard with the increasing dimensionality of the sample space. Another popular approach is the so-called 'quantile approach' \cite{hyndman1996computing}. In this method,  $n$ samples  i.i.d. from known $f(x)$ are sorted in descending order, and the $\big[ \frac{(100-\alpha)}{100} n \big]$-th element of the sorted sample is considered an approximation of $f_{\alpha}$. The main advantage of this approach is that the computational complexity does not increase when the sample space becomes higher-dimensional. 

In the cases where $f(x)$ is unknown and only observable through a set of samples, $f(x)$ can be estimated from the set of samples by, e.g., kernel smoothing \cite{hyndman1996estimating, bashtannyk2001bandwidth, samworth2010asymptotics}. In fact, this is the approach we have adopted in the experiments, based on the implementation of the package \texttt{hdrcde}\footnote{\texttt{https://github.com/robjhyndman/hdrcde}} in R. However, there exists other non-parametric alternatives too, such as the heuristic approach by \citet{waltman2014algorithm}.  

\section{Additional results}

\subsection{Additional results from Section \ref{sec:exp_ho}}
\label{sec:ap_ho_res}

\begin{figure}[!ht]
     \centering
     \begin{subfigure}[b]{0.48\textwidth}
         \centering
         \includegraphics[width=\textwidth]{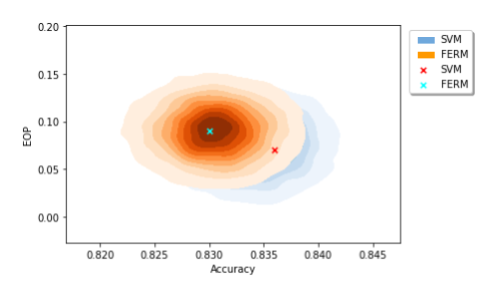}
         \caption{}
         \label{fig:case2_adult_post}
     \end{subfigure}
     \hfill
     \begin{subfigure}[b]{0.48\textwidth}
         \centering
         \includegraphics[width=\textwidth]{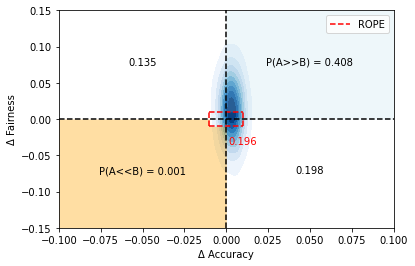}
         \caption{}
         \label{fig:case2_adult_hdr}
     \end{subfigure}
        \caption{(a) The joint posterior distribution of accuracy and equality of opportunity for the methods ($\mathcal{A}$) SVM with a non-linear kernel (blue) and ($\mathcal{B}$) FERM, as well as the classic hold-out result, denoted by the crosses. The overlap between the posterior distributions is significant. (b) The HDR of the 2-dimensional posterior distribution of the difference $\delta$ between the performances of the algorithm. The x-axis denotes the accuracy (re. the difference in accuracy) and the y-axis reflects the fairness guarantees (re. the difference in fairness), in this case, measured by the EOp fairness metric. Although classical results suggest that ($\mathcal{A}$) ourperforms ($\mathcal{B}$) in all the objectives, according to the statistical analysis the probability for such event is only 0.41.}
        \label{fig:case2_adult}
\end{figure}

In this section we provide additional results regarding the comparison between classic and statistical conclusions in the case of the hold out evaluation for the classification task defined by the Adult Income dataset. The distributions regarding \textbf{case study 2} are shown in Figure \ref{fig:case2_adult}. Besides, we consider an additional case study. 

\begin{figure}[!ht]
     \centering
     \begin{subfigure}[b]{0.48\textwidth}
         \centering
         \includegraphics[width=\textwidth]{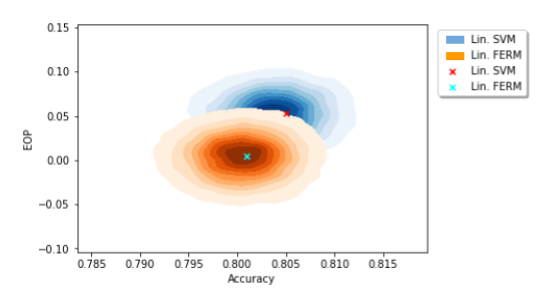}
         \caption{}
         \label{fig:case3_adult_post}
     \end{subfigure}
     \hfill
     \begin{subfigure}[b]{0.48\textwidth}
         \centering
         \includegraphics[width=\textwidth]{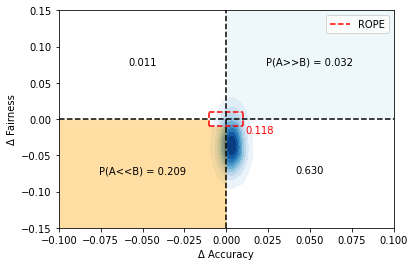}
         \caption{}
         \label{fig:case3_adult_hdr}
     \end{subfigure}
        \caption{(a) The joint posterior distribution of accuracy and equality of opportunity for the methods ($\mathcal{A}$) SVM with a linear kernel (blue) and ($\mathcal{B}$) linear FERM, as well as the classic hold-out result, denoted by the crosses. (b) The 2-dimensional posterior distribution of the difference $\delta$ between the performances of the algorithm. The x-axis denotes the accuracy (re. the difference in accuracy) and the y-axis reflects the fairness guarantees (re. the difference in fairness), in this case, measured by the EOp fairness metric. The classical results suggest that no algorithm outperforms the other in both objectives. However, the probability of total outperforming is not zero: according to the statistical analysis the probability that ($\mathcal{B}$) outperforms ($\mathcal{A}$) in both objectives is 0.21.}
        \label{fig:case3_adult}
\end{figure}

\textbf{Case study 3: ($\mathcal{A}$) Naïve Linear SVM and ($\mathcal{B}$) Linear FERM.} Based on the classic hold-out results, method ($\mathcal{A}$) has a better predictive performance (0.805 vs. 0.801), but provides worse fairness guarantees in terms of EOp (0.05 vs. 0.01). Thus, a conventional evaluation setting suggests that no algorithm outperforms the other in both objectives. However, with regards the statistical analysis, the probability of total outperforming is not zero (see Figure \ref{fig:case3_adult}): actually, there is a probability of 0.21 that algorithm (B) outperforms (A) in both objectives (the probability of the opposite outperforming event is much smaller: 0.03). Nonetheless, it is true that a significant amount of the area of $\delta$ (0.63) is located in the 4th quadrant, which describes the probability that ($\mathcal{A}$) has better predictive accuracy, but ($\mathcal{B}$) provides better fairness guarantees.

\newpage

\subsection{Additional results from Section \ref{sec:exp_cv}}
\label{sec:ap_cv_res}

\begin{figure}[!ht]
     \centering
     \begin{subfigure}[b]{0.48\textwidth}
         \centering
         \includegraphics[width=\textwidth]{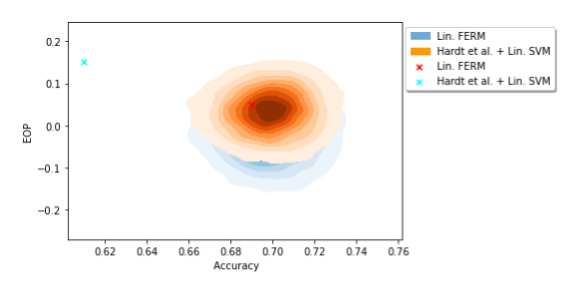}
         \caption{}
         \label{fig:case2_german_post}
     \end{subfigure}
     \hfill
     \begin{subfigure}[b]{0.48\textwidth}
         \centering
         \includegraphics[width=\textwidth]{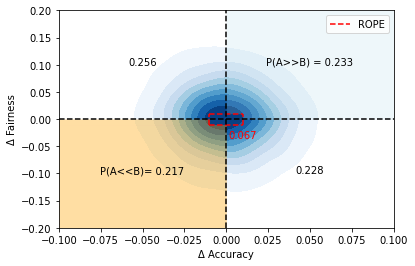}
         \caption{}
         \label{fig:case2_german_hdr}
     \end{subfigure}
        \caption{(a) The joint posterior distribution of accuracy and equality of opportunity for the methods ($\mathcal{A}$) linear FERM (blue) and ($\mathcal{B}$) the combination between SVM with a linear kernel and the post-processing method HardtPP, as well as the classic hold-out result, denoted by the crosses. (b) The 2-dimensional posterior distribution of the difference $\delta$ between the performances of the algorithm. The x-axis denotes the accuracy (re. the difference in accuracy) and the y-axis reflects the fairness guarantees (re. the difference in fairness), in this case, measured by the EOp fairness metric. Although classical results suggest that ($\mathcal{A}$) outperforms ($\mathcal{B}$) in performance and fairness guarantees, according to the UM framework, the probability for such event is only 0.23. What is more, the probability that ($\mathcal{B}$) outperforms ($\mathcal{A}$) in performance and fairness is 0.22.}
        \label{fig:case2_german}
\end{figure}

\begin{figure}[!ht]
     \centering
     \begin{subfigure}[b]{0.48\textwidth}
         \centering
         \includegraphics[width=\textwidth]{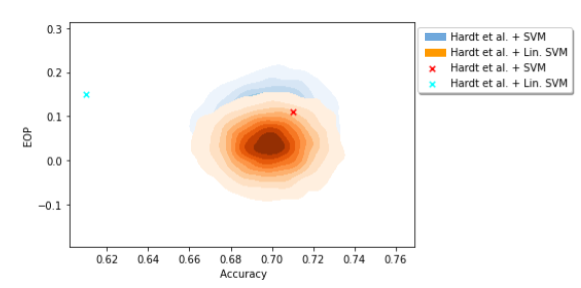}
         \caption{}
         \label{fig:case3_german_post}
     \end{subfigure}
     \hfill
     \begin{subfigure}[b]{0.48\textwidth}
         \centering
         \includegraphics[width=\textwidth]{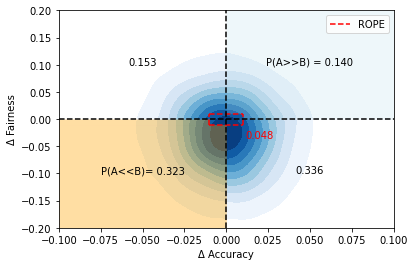}
         \caption{}
         \label{fig:case3_german_hdr}
     \end{subfigure}
        \caption{(a) The joint posterior distribution of accuracy and equality of opportunity for the methods ($\mathcal{A}$) linear SVM plus the post-processing method by HardtPP (blue) and ($\mathcal{B}$) the non-linear SVM plus the post-processing method by HardtPP, as well as the classic hold-out result, denoted by the crosses. (b) The 2-dimensional posterior distribution of the difference $\delta$ between the performances of the algorithm. The x-axis denotes the accuracy (re. the difference in accuracy) and the y-axis reflects the fairness guarantees (re. the difference in fairness), in this case, measured by the EOp fairness metric. Although classical results suggest that ($\mathcal{A}$) outperforms ($\mathcal{B}$) in both objectives, according to the statistical analysis the probability for such event is only 0.144. What is more, actually the probability that ($\mathcal{B}$) outperforms ($\mathcal{A}$) in accuracy and EOp is even higher: 0.322.}
        \label{fig:case3_german}
\end{figure}

In this section we provide additional results regarding the comparison between the classic and UM-based conclusions in the case of the 10-fold CV evaluation framework for the German Credit dataset. In particular, we provide the graphical results corresponding to \textbf{case study 2} and \textbf{case study 3} from Section \ref{sec:exp_cv}, in Figures \ref{fig:case2_german} and \ref{fig:case3_german}, respectively.

\end{document}